\documentclass[lettersize,journal]{IEEEtran}

\usepackage[caption=false,font=normalsize,labelfont=sf,textfont=sf]{subfig}

\usepackage{url}
\usepackage{cite}
\usepackage{tikz}
\usepackage{xcolor}
\usepackage{stfloats}
\usepackage{amsmath,amssymb,amsfonts}
\usepackage{multicol}
\usepackage{tabularx}
\usepackage{steinmetz}
\usepackage{mathtools}
\usepackage{booktabs}
\usepackage{textcomp}
\usepackage{wrapfig}
\usepackage{subfig}
\usepackage{pifont}
\usepackage{verbatim}

\usepackage{multirow}

\usepackage{graphicx}
\graphicspath{ {./images/} }
\usepackage[justification = centering]{caption}

\newcommand*\circled[1]{\tikz[baseline=(char.base)]{
            \node[shape=circle,fill,inner sep=1pt] (char) {\footnotesize \textcolor{white}{#1}};}}
            
\usepackage[colorlinks,
            linkcolor=red,
            anchorcolor=red,
            citecolor=red]{hyperref}  
            
\usepackage{hyperref}
\DeclareUnicodeCharacter{2061}{}
\usepackage{algorithm}
\usepackage{algorithmicx}
\usepackage{algpseudocode}
\usepackage{array}
\floatname{algorithm}{Algorithm}

\begin{document}

\title{Efficient Traffic State Forecasting using Spatio-Temporal Network Dependencies: A Sparse Graph Neural Network Approach}

\author{Bin Lei,~\IEEEmembership{Student Member,~IEEE}, Shaoyi Huang,~\IEEEmembership{Student Member,~IEEE,} 
 Caiwen Ding,~\IEEEmembership{Member,~IEEE}, Monika Filipovska†,~\IEEEmembership{Member,~IEEE}.
 \IEEEcompsocitemizethanks{\IEEEcompsocthanksitem  
 \IEEEcompsocthanksitem †: Corresponding author
 \IEEEcompsocthanksitem  Bin Lei, Shaoyi Huang, Caiwen Ding  are with the  Department  of  Computer  Science  and  Engineering  at the University of Connecticut, Storrs, CT, USA.
 \IEEEcompsocthanksitem Monika Filipovska is with the  Department of Civil and Environmental Engineering  at the University of Connecticut, Storrs, CT, USA.
(E-mail: \{bin.lei, shaoyi.huang, caiwen.ding, monika.filipovska\}@uconn.edu).

 }}



\maketitle

\begin{abstract}
Traffic state prediction in a transportation network is paramount for effective traffic operations and management, as well as informed user and system-level decision-making. However, long-term traffic prediction (beyond 30 minutes into the future) remains challenging in current research. In this work,  we integrate the spatio-temporal dependencies in the transportation network from network modeling, together with the graph convolutional network (GCN) and graph attention network (GAT).  To further tackle the dramatic computation and memory cost caused by the giant model size (i.e., number of weights) caused by multiple cascaded layers, we propose sparse training to  mitigate the training cost, while preserving the prediction accuracy.  It is a process of training using a fixed number of nonzero weights in each layer in each iteration.

We consider the problem of long-term traffic speed forecasting for a real large-scale transportation network data from the California Department of Transportation (Caltrans) Performance Measurement System (PeMS). Experimental results show that the proposed GCN-STGT and GAT-STGT models achieve low prediction errors on short-, mid- and long-term prediction horizons, of 15, 30 and 45 minutes in duration, respectively. Using our sparse training,
we could train from scratch
with high sparsity (e.g., up to 90\%), equivalent to 10$\times$ floating point operations per second (FLOPs) reduction on computational cost using the 
and same epochs as dense training,
and arrive at a model with very small accuracy loss compared with the original dense training.

\end{abstract}

\begin{IEEEkeywords}
Graph Neural Network, Spatio-Temporal, Traffic State Forecasting, Sparse Training, Efficient
\end{IEEEkeywords}

\section{Introduction}
\IEEEPARstart {T}{raffic} state forecasting is a central problem in Intelligent Transportation Systems (ITS) since predictions of the state of traffic in a transportation network is paramount for effective traffic operations and management, as well as informed user and system-level decision-making \cite{luttrell2008predicting}. In recent years, traffic forecasting has been enabled by the increasing quantity and quality of available traffic data. However, accurate modeling and prediction of traffic state, especially across large-scale networks, is a highly nonlinear and complex problem. The network performance is susceptible to fluctuations due to a combination of factors including traffic incidents, weather conditions, work zones, user choices, etc \cite{chen2019operational}. Furthermore, the effects of these factors on traffic spread spatially through the transportation network, and often in a temporally lagged manner as traffic propagates through the network.

 Short-term traffic state prediction approaches are common in the literature, with both statistical and  machine learning methods performing well for 1 to 5-minute prediction horizons \cite{nagy2018survey}. However, longer-term traffic prediction, for example beyond 30 minutes into the future, presents a more difficult problem. This is expected, since it is more difficult to anticipate how traffic patterns will spread spatially and temporally over longer prediction horizons, and the potential effects of a range of measurable and immeasurable external factors become more uncertain \cite{filipovska2021tt}. 
 
 In the literature on traffic state prediction, common early approaches come from time-series analysis \cite{ahmed1979analysis, williams2003modeling, tavana2000estimation}. The immediate limitation of time-series approaches is their singular focus on the temporal component, without accounting for spatial effects. Thus, time-series-based approaches are often applied to individual road segments or small isolated sequences of road segments. However, previous work has shown the importance of incorporating both spatially and temporally lagged features in traffic prediction \cite{filipovska2019, filipovska2020traffic}. Furthermore, with the growing availability of traffic data, machine learning methods are becoming more common and necessary for traffic prediction tasks, due to their ability to capture more complex phenomena. 
 
 This study considers the problem of long-term traffic speed forecasting for a real large-scale transportation network, via an approach that integrates the understanding of spatio-temporal dependencies in the transportation network from network modeling, together with graph-based neural network approaches suitable for large-scale data analysis and complex modeling. Several strategies have been proposed for modeling temporal dynamics and spatial dependencies in the literature, and we focus on two architectures based on graph neural networks (GNNs), the graph convolutional network (GCN) \cite{bruna2014GCN} and graph attention network (GAT) \cite{Velickovic2018GraphAN}. An important consideration for implementing GNN models in real-world large-scale graphs for internet-of-things (IoT) applications is the outsized training computational cost as GNN model sizes continue to grow. Training such large-scale GNNs requires high-end servers with expensive GPUs that are difficult to maintain. A recent study shows that it can take 1600 GPU days to train a GNN model  with 10.7 million parameters \cite{hu2021forcenet}. The computational challenges of training massive GNNs with billions of parameters on large-scale graphs is an important and emerging problem in the machine learning community.
 
 The contribution of this paper is three-fold. (1) It proposes a spatio-temporal GNN-based approach for traffic prediction that used the underlying structure of the transportation network to extract spatio-temporal features traffic speed data throughout the network. (2) To accelerate the GNN training and inference, we use a sparse training method, which converts the computation from dense to sparse by removing large portion of the weights (say, up to 90\%), thus reducing the memory consumption and the total amount of computation with small accuracy loss. (3) The numerical experiments demonstrate the applicability of the proposed approach to a large-scale transportation network and its transferability to time periods not seen during training. 

\section{Background}
\subsection{Statistical Approaches for Traffic Prediction}
Traffic state forecasting problems focus on predicting traffic during a given time interval or prediction horizon in terms of aggregate macroscopic characteristics of traffic flow: speed, density, or flow. Speed is most often used as the quantity of interest in prediction, as a more user-centric measure of traffic state compared to flow or density. Prediction of traffic flow levels is also common as it can offer insights into the level of congestion. Earlier studies use time-series analysis approaches to model the evolution of traffic state over time as a time-series, as introduced by Ahmed and Cook\cite{ahmed1979analysis} and use time-series forecasting methods for prediction such as ARIMA \cite{williams2003modeling,van1996combining,tavana2000estimation} and Kalman filtering methods \cite{whittaker1997tracking}. These approaches initially focused on univariate prediction at a given location, and were later extended to multi-variate problems that aim to simultaneously capture traffic state at various locations with approaches such as space-time ARIMA (STARIMA) \cite{stathopoulos2003multivariate, kamarianakis2003forecasting}, and combined with adaptive Kalman filtering \cite{guo2014adaptive}.

\subsection{Machine Learning for Traffic Prediction}
With the increase of data availability in transportation systems, data-driven machine learning approaches for traffic forecasting have become more common in recent years. Many of the initial studies focus on short term prediction, such as \cite{hosseini2012short} using multi-layer perceptron models, as the simplest form of neural networks, \cite{jia2016traffic, huang2014deep} employing deep belief networks. Wu and Tan \cite{wu2016short} present an approach for short-term traffic flow forecasting with spatio-temporal correlation in a hybrid deep-learning framework combining convolutional neural networks (CNN) and long short-term memory (LSTM) networks. Tahlyan et al.\cite{tahlyan2021meta} present a meta-learner ensemble framework for real-time short-term traffic speed forecasting. Some recent studies also focus on longer-term prediction, such as \cite{lv2014traffic} presenting a deep learning stacked auto-encoder approach for a prediction horizon of up to 60 minutes, and \cite{ma2015long} suing an LSTM neural network to capture long-term temporal dependence. 
 
 While all of the studies referenced above consider traffic prediction from a primarily temporal aspect at isolated locations, a few recent studies consider the problem of spatio-temporal traffic state prediction. Li et al.\cite{li2017diffusion} propose a diffusion convolutional recurrent neural network (RNN) to capture spatial dependency in traffic state predictions. Yu et al. \cite{yu2017spatiotemporal} consider the problem of large-scale network-wide traffic speed prediction and implement a spatiotemporal recurrent convolutional network (SRCN) approach for short term prediction. Ma et al. \cite{ma2017learning} approach a similar problem using a convolutional neural-network (CNN) model aiming to capture spatiotemporal traffic dynamics, which was shown to be well suited for short to medium-term prediction of 10 to 20 minutes. 
 
 The recent emergence of GNNs has helped advance the methods for spatio-temporal traffic forecasting across networks. For mid- and long-term prediction tasks across space, \cite{yu2017spatio} introduce a graph convolutional network (GCN), referred to as spatio-temporal GCN (STGCN), while \cite{zhang2019spatial} and \cite{kong2020stgat} both present a spatial-temporal graph attention network (STGAT) approach. The key difference between the STGCN and STGAT approaches is the use of a graph convolutional network (GCN) in the former and a graph attention network (GAT) in the latter. GAT modifies the convolution operation in a GCN with an attention mechanism whose purpose is to perform node classification of graph-structured data by computing hidden representations of each node in the graph by attending over the node’s neighbors. Thus, GAT performs the functions of a GCN while attuning itself to the particular features that are most important in the network. GAT also promises efficient (parallelizable) operations, application to graph nodes with different degrees, and applicability to inductive learning problems including models that are generalizable to previously unseen graphs. Details on the GAT architecture can be found in the original work \cite{velivckovic2017graph}. 
 
 An important limitation of the spatio-temporal models in recent studies \cite{yu2017spatio,zhang2019spatial,kong2020stgat} is that they do not take into account the underlying transportation road network. Instead, they construct artificial distance-based adjacency matrices using Euclidean distance to infer the connections between locations and generate the graphs. Thus, the spatial component does not account for the structure of the transportation network and the actual flow of traffic over space, but is simply based on the spatial proximity of the locations where data is observed.
 
\subsection{Weight Sparsification in Graph Neural Network}
State-of-the-art deep neural networks have large model size and computational requirements, which limits their ability to provide a user-friendly experience \cite{he2016deep, gpt3,huang2022sparse,huang2022automatic,huang2021hmc}. To address the challenges, weight sparsification has been investigated.  The key idea is to represent a neural network with a much simpler model (set a group of weight values to be zero), therefore bringing acceleration in computation and reduction in memory~\cite{chen2021re,peng2021accelerating,peng2022length,qi2021accelerating,qi2021accommodating,peng2022towards,manu2021co}. Recent work by Chen \cite{Chen2021AUL} proposes a weight sparcification framework for GNNs, called Unified GNN Sparsification (UGS), that pruned the input graph as well as the model weights using the well-studied \textit{lottery ticket hypothesis} weight pruning method \cite{frankle2018lottery}. However, these work mainly focus on generating on the inference efficiency, and often need to use more training epoches on training.


\section{Problem definition}
The problem considered in this paper is as follows. Given the observations of aggregate traffic speeds across a number of spatially distributed road segments in a transportation network for a historical time horizon $H$, the goal is to predict the aggregate traffic speed for the same locations over a future time horizon $T$. The road segments at which traffic speeds are observed are connected to one another to form the transportation network on which traffic flows between these spatially distributed locations. 
 Towards this goal, we define the underlying transportation network as a graph $G(N,A,W)$ where $N$ is the set of nodes (i.e., vertices), A are the set of links (i.e., edges) connecting those vertices and $W \in \mathbb{R}^{|N| \times |N|}$ is a weighted adjacency matrix for network. Typical representations of transportation networks model the intersections as nodes connected by road segments which are modeled as links in the graph. However, in defining this problem, there is a need to model the traffic monitoring stations (i.e. sensors at fixed locations) on road segments as the set of nodes $N$ and use the links $A$ to model the connections between those road segments. Therefore, $G(N,A,W)$ is a transformed version of the road network, so that the nodes $i \in N$ correspond to the $|N|$ observation locations, i.e., the stations yielding traffic speed measurements and the links $(i,j) \in A$ are constructed where there are direct station-to-station connections on the road network. Thus, if $(i,j) \in A$ then the stations $i$ and $j$ are consecutive or connected by a consecutive sequence of road segments on which traffic flows directly from $i$ to $j$, with no intermediate stations along the way. The weight matrix supplies additional information regarding the connections between nodes. Firstly, let $w_{ij} \in W$ denote the $i$, $j$ element of the $|N|\times |N|$ matrix, and

\begin{equation}
\label{eq:1}
w_{ij} = 
\begin{cases}
w(i,j) & if(i,j) \in A\\
0 & otherwise
\end{cases}
\end{equation}
where function $w\text{:}(i,j)\in A \rightarrow \mathbb{R}$ is a measure of closeness between the connected nodes $i$ and $j$. The measure of closeness is often inversely related to distance denoting the relatedness between observations at nodes $i$ and $j$.

 To define this problem mathematically, suppose that time is discretized into short fixed-duration intervals $\delta$, which will be referred to as time steps, so that the historical horizon $H$ consists of $|H|$ time steps and the horizon $T$ consists of $|T|$ time steps. Suppose $N$, as defined above, is the set of locations where traffic is observed, numbering $|N|$ locations. Then an observation of traffic speeds for a single time step $t$ is a vector of size $|N|$ denoted $\boldsymbol{V_{t}} =[v_t^0,...,v_t^{|N|}] \in \mathbb{R}^{|N|}$ where
$v_t^i \in \mathbb{R}$ is a scalar value for the speed at location $i \in N$. Thus, at a current time $t$ the goal is to find $\hat{\boldsymbol{V}}_{t+1},\hat{\boldsymbol{V}}_{t+2},...\hat{\boldsymbol{V}}_{t+|T|}$ based on the observed $\hat{\boldsymbol{V}}_{t-|H|+1},...\hat{\boldsymbol{V}}_{t-1},\hat{\boldsymbol{V}}_{t}$ to maximize:
\begin{equation}
\label{eq:2}
{log P(\hat{\boldsymbol{V}}_{t+1},
...\hat{\boldsymbol{V}}_{t+|T|}| \\  \hat{\boldsymbol{V}}_{t+1-|H|},...,\hat{\boldsymbol{V}_{t-1}},\hat{\boldsymbol{V}_{t})}}
\end{equation}

where $P(\cdot)$ denotes the unknown, dynamic, data-generating process. 

\section{Methods}
\subsection{Preliminaries}
As seen above, traffic state prediction can be seen as a time-series forecasting problem, where the traffic state in future time intervals is predicted based on observations during past time-intervals. However, traffic in the transportation network evolves both over space and time, and temporal patterns are not independent of spatial dependencies due to the flow of traffic across the network. This paper presents a deep learning approach that predicts traffic state across a transportation network and over medium to long-term prediction horizons. The proposed solution approach should be able to capture and employ knowledge of the spatio-temporal dependencies in traffic state, thus it uses the graph representation of the underlying transportation network as the basis for the GNN-bases model. The transportation network is transformed into a graph, as described in the previous section, with a weighted adjacency matrix determined via Equation \ref{eq:1} where the weights are computed as $w(i,j)=e^{-\omega d_{ij}}$ where $d_{ij}$ is the road-network distance between stations $i,j\in N$ and $\omega$ is a scaling weight factor.


\subsection{Proposed Model Architecture}
\begin{figure}
\centering
\includegraphics[scale=0.4]{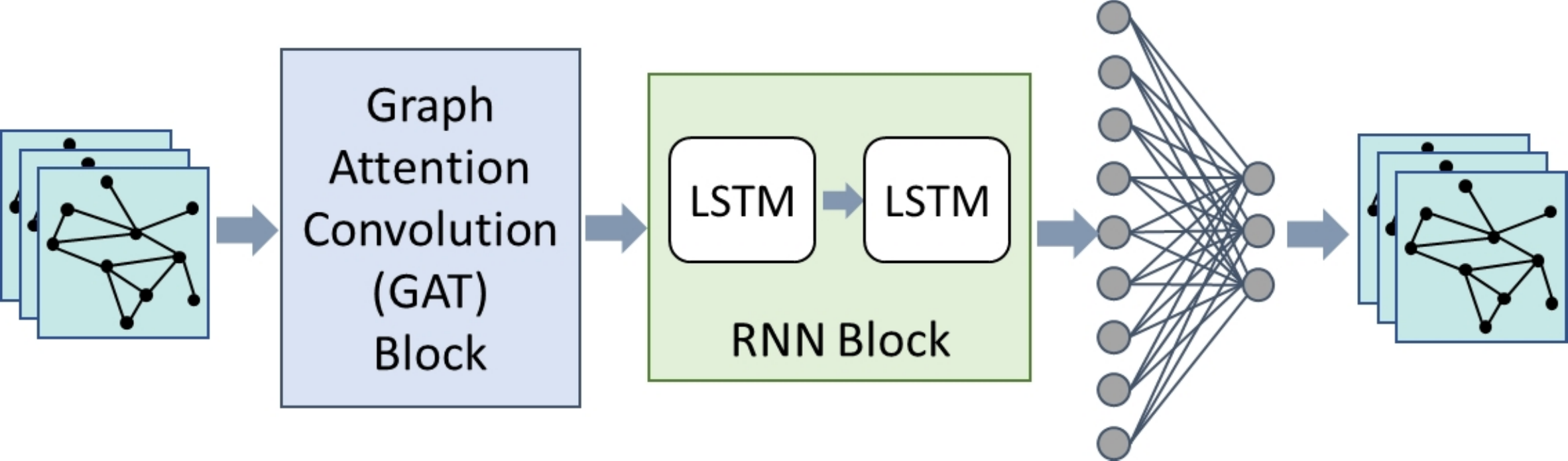}
\caption{Spatio-Temporal Graph Convolutional Network Model Architecture}     \label{Model Architecture}
\vspace{-4mm}
\end{figure}

The proposed spatio-temporal GNN-based traffic prediction approach (STGT) combines neural network components intended to capture spatial and temporal features and patterns in traffic data. The input to the model is the multi-dimensional time-series of the traffic speeds observed over the past $|H|$ time intervals in a historical time horizon across $|N|$ observation locations, denoted $V_t=[v_t^0,...,v_t^{|N|}]\in R^{|N|}$. This input feeds into a spatial GNN-based block that extracts and learns the spatial features. We implement two versions of the model, where the spatial GNN-based component can be a GCN or GAT block, which will be referred to as GCN-STGT and GAT-STGT, respectively. This is followed by an RNN block using a two-layer long short-term memory (LSTM) network to preform temporal feature extraction. Finally, the RNN outputs passes through a fully-connected network to generate the final predictions. 
The model architecture is shown in Figure \ref{Model Architecture}, and the details of the model are described below. 

\subsubsection{Spatio-temporal Representation of Traffic Data}
While GNN models are able to leverage node features to capture spatial dependency of a graph, there are demonstrated challenges in employing them with traffic data. Specifically, Zhang et al. \cite{zhang2019spatial} show that it can be difficult to find proper feature representation of time-series data over space and propose a Speed-to-Vector (Speed2Vec) data embedding mechanism to define feasible feature representations of the time-series data to be used within GNNs. The present study adopts the Speed2Vec representation, where the speed observations at a given node in a fixed-duration historical time horizon are taken as the hidden feature at a time step, embedded in a vector $h_t=[v_{t-F+1},v_{t-F+2},…,v_t]$ where $h_t \in \mathbb{R}^F$ and $F$ is the duration of the historical time horizon in terms of number of time steps. Then, these vectors are used to create the network-wide input $H_S^N$ to the spatial GAT block, such that: 
\begin{equation}
\label{eq:3}
H_S^N = 
\begin{bmatrix} h_1^1 &\cdots& h_S^1 \\
\vdots & \ddots & \vdots \\
h_1^{|N|} & \cdots &h_S^{|N|} \end{bmatrix}
\end{equation}
where $H_S^{N} \in \mathbb{R}^{S\times{|N|}\times F}$, $S$ is the length of the temporal sequence, and $|N|$ is the number of nodes (i.e., stations generating traffic observations). The historical horizon duration $F$ can vary depending on the specific dataset and application of the approach. For example, if 1-minute observations are collected, the historical horizon might contain a larger number of time-steps compared to cases when 5-minute data is available. Similarly, the historical horizon might need to be adjusted depending on the desired prediction horizon $T$, as specified in the problem definition.

\subsubsection{Spatial Feature Extraction} 
The proposed model architecture is tested with two types of GNNs for spatial feature extraction: a GCN and a GAT block. The architecture is implemented according to the original GAT implementation by \cite{Velickovic2018GraphAN} where the GAT version employs a multi-head attention mechanism to enable the model to jointly learn spatial dependencies through multiple attention blocks. The GAT uses the provided graph structure, in this case the transportation road network, and performs self-attention over the nodes $i\in N$ to compute the so-called attention coefficients $e_ij$ for each of its neighboring nodes $j\in N$ s.t. $(i,j)\in A$, which indicate the importance of node $j$'s features to node $i$. Therefore, the GAT learns a weight matrix for the relatedness between the graph nodes. On the other hand, the GCN architecture is simpler, in that the weight matrix is assumed to be as provided to the model, in this case the inverse-distance-based weight matrix defined previously. Comparing the performance of GCN-STGT and GAT-STGT model architectures will allow for an understanding how much of the attention information can be captured by the interpretable inverse-distance-based weight matrix in the GCN-STGT, and how much improvement can be brought on by the attention mechanism employed in GAT-STGT. 

\subsubsection{Temporal Feature Extraction}  
The traffic state data, with its distinct time-series format is suitable for use with recurrent neural networks (RNNs) which can be leveraged to learn temporal dependencies for time-series prediction. In particular, LSTM networks are the most commonly used RNNs, especially when long-term dependencies need to be captured \cite{yu2019review}. The LSTM architecture uses gating units and cell states for the flow of information for long-term time series prediction. While the cell states store memory information and pass through all the time iterations, the gating units are used to decide whether to add or remove information for a cell state. Additionally, LSTM uses so-called forget gates which decide which information can be removed from a cell state. More details and the mathematical expressions of an LSTM with a forget gate are presented in the original work by Gers et al. \cite{gers2000learning}. 

\subsection{Sparse Model Structure}
\begin{figure*}[htpb!]
\centering
\captionsetup{justification=centering}
 \includegraphics[width = 0.99\linewidth]{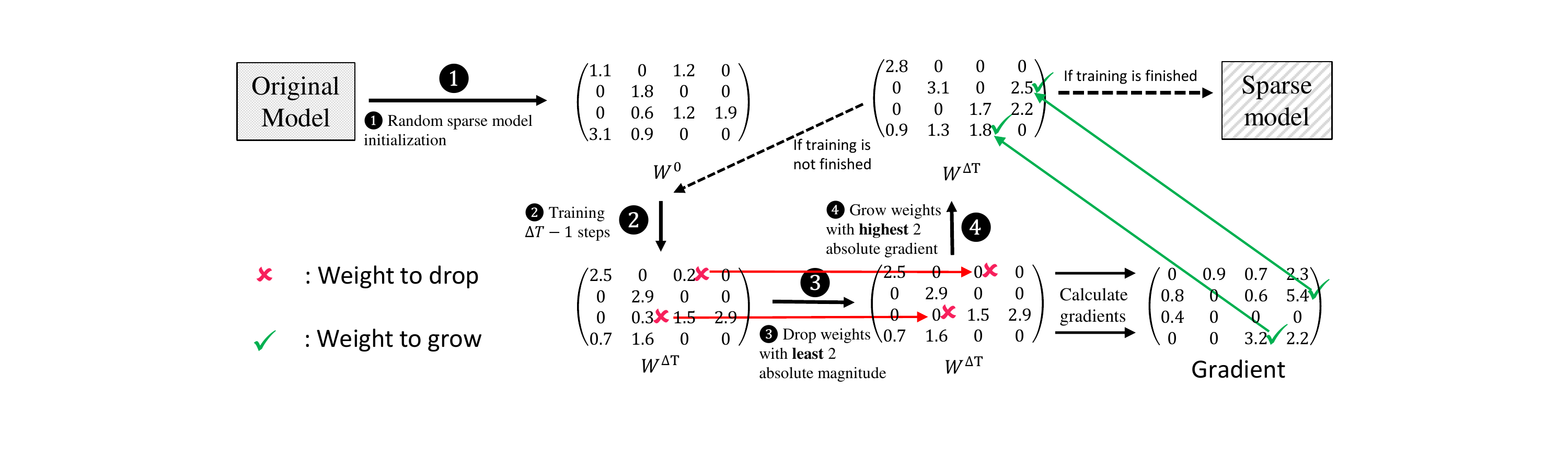}
 \centering
     \caption{Iterative drop \& grow based sparse training process.}
     \vspace{-4mm}
     \label{fig:sparetraining}
\end{figure*}

To reduce the amount of arithmetic operations to be performed and the number of weights to be stored on GCN-STGT  and GAT-STGT, we will sparsify our model by removing some of the neuronal connections. In general, sparsifying the model can reduce the computing time significantly with almost no loss of accuracy.\cite{hoefler2021sparsity}
The details of the approach are shown in Figure \ref{fig:sparetraining}, where first a random sparse model is initialized, then at regular intervals $n$ weights with the lowest absolute value of gradient are dropped, while $n$ weights with the highest value are grown.

Dynamic sparse training is the process of training with fixed number of nonzero weights in each neural network layer. We use a toy example to illustrate the sparse training dataflow. For simplicity, we use the matrix with a size of $4 \times 4$ to represent a weight tensor in the neural network. The sparse training is comprised of 4 steps as follows. 

\begin{itemize}
\item \protect\circled{1} The weight tensor is random sparsified as $W^0$ at a given sparsity $S = 0.5$, which means 50\% of weights will be deactivated (set as zeros) and others remain activate (non-zero). 
\item \protect\circled{2} The sparsified tensor will be trained $\Delta T - 1$ iterations, where $\Delta T$ is the drop-and-grow frequency. During the $\Delta T - 1$ epochs, the non-zero elements in weight tensor are updated following the standard training process, while the zero elements will remain as zero. At the $i$-th iteration, the weight tensor is denoted as $W^i$, while the gradient is denoted as $G^i$. 
\item \protect\circled{3} At the $\Delta T$-th iteration, we first drop $k$ weights that are closed to zero or set the weights that have the least $k$ absolute magnitude as zeros ($k = 2$). Then, 
\item \protect\circled{4} we grow the weights with the highest $k$ absolute gradients back to nonzero (updating the weights with the highest $k$ absolute gradients to nonzero in the following weights updating iteration). During the process, the number of activated weights are kept the same, i.e., the newly activated (non-zero) weights are the same amount as the previously deactivated (zero) weights. 
\end{itemize}

\protect\circled{2}\protect\circled{3}\protect\circled{4} will be repeated until the end of the training.

See Algorithm \ref{alg:alg1} for detailed algorithm.
\begin{algorithm}
\small
\caption{Sparse training algorithm}\label{alg:alg1}
\begin{algorithmic}[1]
\Require Weight of each layer $W^l$, Sparse matrix of each layer $M^l$, The number of data $N$, Update frequency $f$, Drop rate $k$, Sparsity rate $d$
\Ensure Updated sparse matrix $M_{new}^l$, Updated weight matrix $W_{new}^l$

\If {$N ==0$} 
 \State $M^l \gets Initialize()$
\EndIf
\If {$N \% f ==0$} 
\For{$\textit{each layer}\in model$} 
\State $W^l_{Sorted} \gets Sort(W^l.view(-1))$ 
\For{$M^l_{ij}\in M^l$}
\If{$W^l_{ij} \ge W^l_{Sorted}[len(W^l_{Sorted})\times k]$}
\State $M^l_{ij}\gets 1$
\Else
\State $M^l_{ij}\gets 0$
\EndIf
\EndFor
\EndFor
\For{$\textit{each layer}\in model$} 
\State $g^{W^l} \gets W^l.grad.clone() $ 
\State $g^{W^l}_{Sorted} \gets Sort(g^{W^l}.view(-1))$ 
\State $g^{W^l}_{Sorted} \gets g^{W^l}_{Sorted}\bigodot M^l$ 
\For{$M^l_{ij}\in M^l$}
\If{$g^{W^l} \ge g^{W^l}_{Sorted}[len(W^l_{Sorted})\times (1-k)]$}
\State $M^l_{ij}\gets 1$
\EndIf
\EndFor
\EndFor
\For{$\textit{each layer}\in model$} 
\State $W^l\gets W^l\bigodot M^l$
\EndFor
\EndIf
\State \Return{$W^l, M^l$}
\end{algorithmic}
\end{algorithm}

\section{Experiments}

\subsection{Data Description}
\begin{figure}
\centering
\includegraphics[scale=0.5]{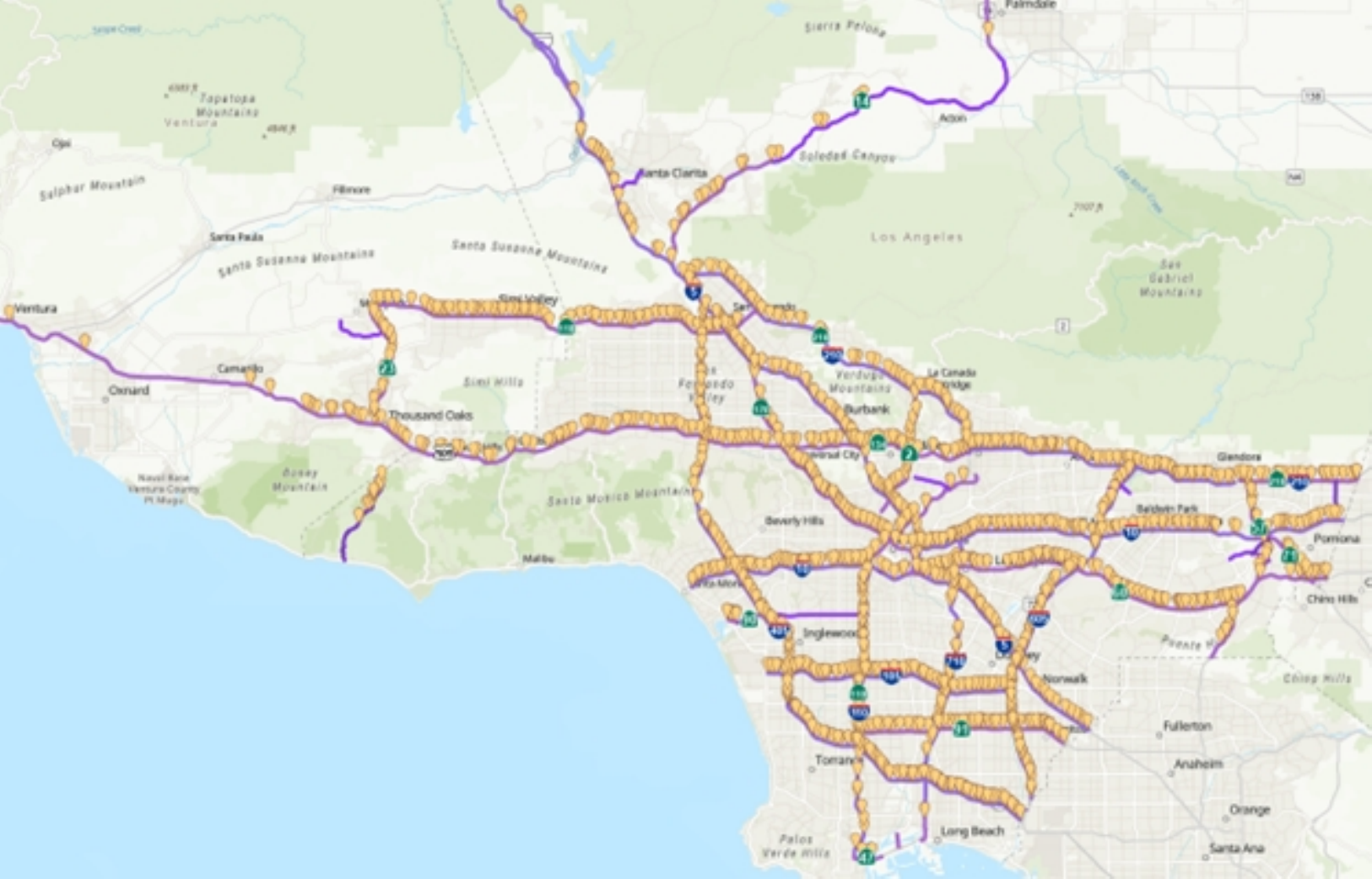}
\caption{Locations of 2471 working sensor stations in district on the corresponding road network }     \label{Locations of 2471}
\end{figure}

The numerical experiments use data from the California Department of Transportation (Caltrans) Performance Measurement System (PeMS), which collects real-time information from nearly 40,000 sensor stations across the freeway system of California and provides an Archived Data User Service (ADUS) containing over 10 years of data for historical analysis. 
Traffic state prediction in this study is performed using the 5-minute station data, which include aggregate information regarding the traffic speed, flow, and occupancy on road segments at sensor locations. Specifically, we select the stations located in District 7, encompassing the Los Angeles metropolitan area, shown on the map in Fig. \ref{Locations of 2471}. 
 
 The primary data for this analysis were the 5-minute data from district 7 stations for May and June of 2019. Data for six additional months were also obtained to test the transferability of the model across time periods when significant changes in demand and traffic patterns may have occurred. The station metadata for district 7 were used to obtain the location, corresponding road and direction of travel for each sensor station. This information was combined with the California Road System (CRS) web map and functional classification of roads. 
 
 The original data contained information from a total of 4,854 sensor stations within the area of district 7. In the pre-processing stage, two types of missing data patterns were found: (1) some days had no observations (or had only nan value observations) for all or most stations, and (2) certain stations had missing data for multiple full days in the specified time period. The data were cleaned by first removing all days where the information for more than half of the stations was missing and then keeping only the set of working stations across the remaining days. After processing, 2,471 working stations remained to be used across all of the 8 months, with a few days of data removed for each month as needed. The locations of these working stations are represented with the yellow pins on the map in Figure \ref{Locations of 2471}.
 
 The process for combining the PeMS Station Metadata with the CRS data to create a graph structure based on the underlying network structure are outlined briefly, following the approach described in the problem definition. First, a set of nodes were created at all road intersections to capture the connectedness of the road network. Second, detectors were matched to the nearest location on the corresponding road, and the graph nodes $N$ were created at these locations. Third, a set of graph edges $A$ were created along the roads and across intersections to connect any two adjacent detector locations. The roads connecting the stations are shown as purple lines in Figure \ref{Locations of 2471}, where only relevant roads containing at least one sensor station are displayed. 
This information was used to generate a weighted matrix $\textbf{A}$, where for each pair of nodes as described previously. The weights were computed using distances along the network roads (i.e., not Euclidean or aerial distances), for the adjacent roads only.

\subsection{Experiment Design}
The numerical experiments are designed for training, validation, and testing of the proposed STGT approach using 60 minutes of historical speed data from all 2,471 working sensors as input for the prediction of the traffic speed for the next 15 to 45 minutes across the same stations. Specifically, the models of the two types, GCN-STGT and GAT-STGT were trained for short-, mid- and long-term prediction of 15, 30 and 45-minutes in duration, respectively. The models are trained on data from the 53 available days in May and June of 2019, which will be referred to as the training time period 0, abbreviated TP0. Training is performed across 200 epochs with validation, and then test on a separate subset of the data not used for training or validation.

 We conduct our GNN sparse training on an Intel Xeon Gold 5218 machine at 2.30 GHz with Ubuntu 18.04 using an Nvidia Quadro RTX 6000 GPU with 24 GB GPU memory. We use python 3.10.8, CUDA 11.6, pytorch 1.12.1 to setup the code.
 We set the total number of epochs to 200 and the batch size equals to 242. For sparse training: The model initialization sparse method is \textit{Erd˝os-R´enyi-Kernel} (ERK) method\cite{evci2020rigging}. We set the death rate as 0.5 and drop-and-grow frequency as 1000 batch iterations. We set the model sparsity as 0.025, 0.05, 0.1, 0.125, 0.25, 0.375, 0.5, 0.625, 0.75, 0.875, 0.9, 0.95, 0.975 to evaluate the performance of the sparse training method and the float point operations (FLOPS) compared with dense model. The performance of the approach is measured using Mean Absolute Errors (MAE), Mean Absolute Percentage Errors (MAPE), and Root Mean Squared Errors (RMSE). 

 To further evaluate the performance of the models on unseen and potentially different traffic patterns, the models are tested via zero-shot transfer with data from different time periods. A total of 3 testing scenarios (time periods) are created: 
\begin{itemize}
\item Testing time period 1 (TP1) during July and August 2019, immediately following the training period. Shifts in traffic patterns are possible in this period due to changes in lifestyle and travel needs during the summer. 
\item Testing time period 2 (TP2) during December 2019 and January 2020, i.e. the following winter season. Though weather changes may not be significant in southern California, shifts in traffic patterns could occur due to seasonal shifts in travel demand. 
\item Testing time period 3 (TP3) during May and June 2020, exactly a year after TP0 which would be expected to be most similar to TP0 due to the typical seasonality in transportation data. However, some important changes could have occurred due to the impacts of the COVID-19 pandemic causing shifts in activity and travel patterns. 
\end{itemize}

\subsection{Evaluation Results}

The primary results from the numerical experiments are shown in Table \ref{tbl:table1} and \ref{tbl:table2}, where each table includes the error measures for the GCN-STGT and GAT-STGT with three different prediction horizons and across all four time periods described above, for the original models and with 90\% sparsity level, respectively. Overall, comparing the error values between GCN-STGT and GAT-STGT, it can be observed that the use of GAT helps reduce the MAPE by 1 percentage point in most cases. This indicates that the inverse-distance based weights derived from the road network are able to capture some of the spatial dependencies and allow for a relatively well performing model, however the self-attuning mechanism can help further improve its performance. Comparing the values in Tables \ref{tbl:table1} and \ref{tbl:table2}, we observe very small increase in error values, indicating the the models' accuracy is preserved while introducing a significant level of sparsity. For the sake of brevity, the error measurement with respect to sparsity are showed for only two cases, the GCN-STGT and GAT-STGT models for a 45-minute prediction horizon, in Figures \ref{GCN 45min} and \ref{GAT 45min}, respectively. 

\subsubsection{Temporal Prediction Performance}
In Table \ref{tbl:table1}, before introducing the model sparsity, each model's performance can be compared across the 3 prediction horizons for a given dataset. For TP0, the GCN-STGT prediction error is highest for the 45-minute horizon, however the performance is not significantly different between the three prediction horizons. With the GAT-STGT approach, the prediction errors are lower than those achieved by GCN-STGT, and the differences in the prediction error for the horizons in the range from 15 to 45 minutes is no longer significant. These results indicate that the proposed approach is suitable for both mid- and long-term traffic prediction. 

Considering the zero-shot transfer tests with the datasets from different prediction horizons, some interesting observations can be made. As expected, errors are lowest when testing on a subset of the training dataset within the original time period (TP0), with MAPE averaging close to 4\% with GCN-STGT and 3\% with GAT-STGT. However, the models' transferability differs between three testing time periods. Namely, the MAPE values for TP3 are consistently the lowest, averaging close to 10\% with GCN-STGT and 9\% with GAT-STGT. This indicates that zero-shot transfer is possible for TP3. However, the MAPE values for TP2 are close to 22\% with GCN-STGT and 21\% with GAT-STGT, while for TP1 the highest error values are observed, with MAPE near 24\% for both model types. This indicates that the models may need to retrained for TP1 and TP2. This lack of transferrability over time, especially for the nearest time period (TP1), reinforces the need for a model that can be trained efficiently, which in this case is made possible by the sparsity techniques discussed in the following section.

\subsubsection{Sparse Training Accuracy Evaluation}
Tables \ref{tbl:table1} and \ref{tbl:table2} show the comparison of GCN-STGT and GAT-STGT on TP1, TP2 and TP3 for three different time steps (15min, 30min, 45min) with 0\% sparsity and with the sparsity level at 90\%, respectively. We use three error measurements: MAE, RMSE and MAPE. The test results on the three datasets show that, for the most cases, the average error of the GAT-STGT method is lower than that of the GCN-STGT method.
The difference between the two approaches is more significant under the 30min and 15min predictions. 

Using typical training time (total training epochs is 200), there is almost no accuracy loss compared to the dense model even at sparsity of 90\% on all three different datasets. For the GCN-STGT method, the error increases slightly until the sparsity reaches 85\%. For the GAT-STGT method, the accuracy-sparsity curve has the Occam’s Hill \cite{rasmussen2000occam} property where the accuracy first increases with increasing sparsity and then decreases. Therefore, the embedding sparsification is favorable for this specific task. 

\begin{table*}[htbp]
  \caption{Comparison of the error values of GCN-STGT and GAT-STGT for different prediction horizons with 0\% sparsity}\label{tab:table1}
  \centering
  \setlength{\tabcolsep}{5.7mm}{
  \begin{tabular}{|c|c|ccc|ccc|}
    \hline
    \multirow{2}{*}{\centering{Dataset}}
    &\multirow{2}{*}{\centering{Error}}
    &\multicolumn{3}{c}{GCN-STGT}
    &\multicolumn{3}{|c|}{GAT-STGT} \\
     \cline{3-8}
     {} &{} & 45min & 30min & 15min & 45min & 30min & 15min  \\\hline
    \multirow{3}{*}{\centering{Training (TP0)}}  
    & MAE & 2.053 &  1.945 &  1.988 & 1.829 & 1.870 & 1.951 
     \\\cline{2-8}    
    &RMSE & 4.325 & 3.77 & 3.897 & 3.657 & 3.861 & 3.803
    \\\cline{2-8}
   & MAPE & 4.194 & 3.350 & 3.407 & 3.121 & 3.379 & 3.319  \
   \\
   \hline
  \multirow{3}{*}{\centering{Time period 1 }}  
  & MAE & 8.051 &  10.67 &  9.515 & 8.082 & 7.585 & 7.468  
  \\\cline{2-8}    
    &RMSE & 11.95 &  12.78 &  13.21 & 10.46 & 11.76 & 11.69  
    \\\cline{2-8}
   & MAPE & 24.73 &  24.02 &  24.91 & 24.58 & 24.16 & 23.85  \
   \\
   \hline  
     \multirow{3}{*}{\centering{Time period 2}}  
     & MAE & 7.410 & 9.618 & 8.687 & 7.280 & 6.910 & 6.798  
     \\\cline{2-8}    
    &RMSE & 11.27 & 11.56 & 12.39 & 10.63 & 10.87 & 10.83
    \\\cline{2-8}
   & MAPE & 22.03 & 21.00 & 22.11 & 21.76 & 21.28 & 21.12  \
   \\
   \hline
     \multirow{3}{*}{\centering{Time period 3}}  
     & MAE & 4.246 & 5.450 & 5.091 & 4.056 & 4.043 & 4.053
     \\\cline{2-8}    
    &RMSE & 7.031 & 7.698 & 7.484 & 6.249 & 6.753 & 6.653
    \\\cline{2-8}
   & MAPE & 9.978 & 10.01 & 10.48 & 9.273 & 9.250 & 9.216  \
   \\
   \hline
  \end{tabular}}
  \label{tbl:table1}
\end{table*}

\begin{table*}[htbp]
  \caption{Comparison of the error values of GCN-STGT and GAT-STGT for different prediction horizons with 90\% sparsity}\label{tab:table2}
  \centering
  \setlength{\tabcolsep}{5.7mm}{
  \begin{tabular}{|c|c|ccc|ccc|}
    \hline
    \multirow{2}{*}{\centering{Dataset}}
    &\multirow{2}{*}{\centering{Error}}
    &\multicolumn{3}{c}{GCN-STGT}
    &\multicolumn{3}{|c|}{GAT-STGT} \\
     \cline{3-8}
     {} &{} & 45min & 30min & 15min & 45min & 30min & 15min  \\\hline
    \multirow{3}{*}{\centering{Training (TP0)}}  & MAE & 2.786 &  2.664 &  2.318 &  3.189 & 2.500 &  2.146 
     \\\cline{2-8}    
    &RMSE & 4.850 & 4.776 & 4.575 & 5.404 & 4.616 & 4.621
    \\\cline{2-8}
   & MAPE & 5.647 & 5.418 & 4.938 & 6.319 & 5.247 & 4.787  \
   \\
   \hline
  \multirow{3}{*}{\centering{Time period 1 }}  & MAE & 8.273 &  10.07 &  9.947 & 8.246 &  8.463 &  8.608  
  \\\cline{2-8}    
    &RMSE & 12.14 &  14.85 &  13.85 &  12.589 & 12.26 &  12.36  
    \\\cline{2-8}
   & MAPE & 24.51 &  26.25 &  25.41 &  24.87 & 25.10 &  24.88  \
   \\
   \hline  
     \multirow{3}{*}{\centering{Time period 2}}  & MAE & 7.477 & 10.25 & 9.214 & 7.468 & 7.633 & 7.873  
     \\\cline{2-8}    
    &RMSE & 11.24 & 13.89 & 12.93 & 11.10 & 11.29 & 11.44
    \\\cline{2-8}
   & MAPE & 21.71 & 23.67 & 22.88 & 21.71 & 22.15 & 22.09  \
   \\
   \hline
     \multirow{3}{*}{\centering{Time period 3}}  & MAE & 4.552 & 6.024 & 5.471 & 4.563 & 4.642 & 4.789
     \\\cline{2-8}    
    &RMSE & 6.954 & 8.054 & 7.843 & 6.956 & 7.040 & 7.130
    \\\cline{2-8}
   & MAPE & 9.900 & 12.19 & 11.03 & 9.921 & 10.09 & 10.23  \
   \\
   \hline
  \end{tabular}}
  \label{tbl:table2}
\end{table*}

\begin{figure*}[htbp]
\centering
\includegraphics[scale=0.5]{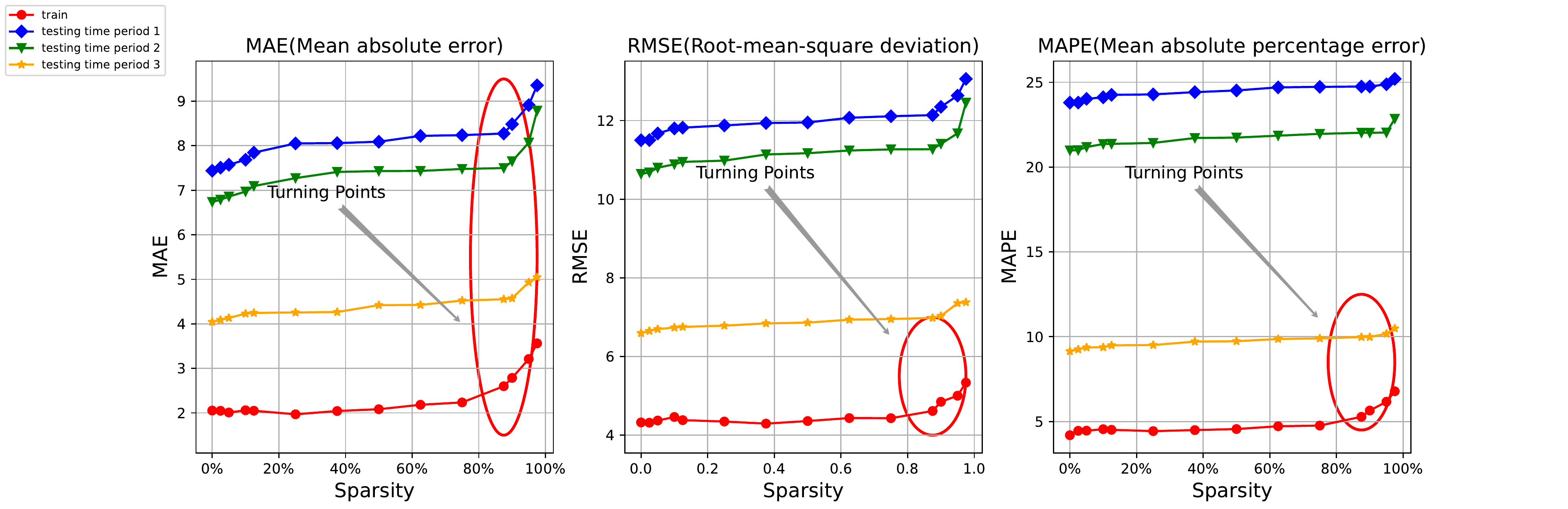}
\caption{Mean absolute error(MAE), Root-mean-square(RMS), Mean absolute percentage error(MAPE) vs. sparsity for testing time period 1, testing time period 2, testing time period 3, using GCN-STGT with a 45-minute prediction horizon} \label{GCN 45min}
\end{figure*}

\begin{figure*}[htbp]
\centering
\includegraphics[scale=0.5]{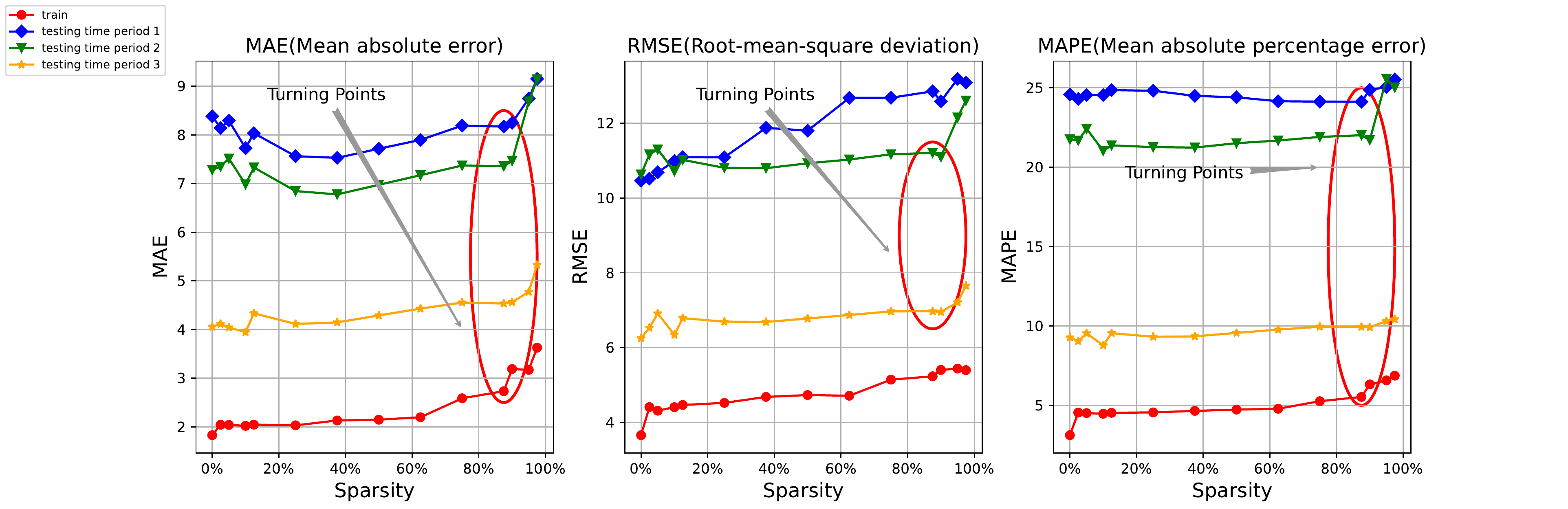}
\caption{Mean absolute error(MAE), Root-mean-square(RMS), Mean absolute percentage error(MAPE) vs. sparsity for testing time period 1, testing time period 2, testing time period 3, using GAT-STGT with a 45-minute prediction horizon} \label{GAT 45min}
\end{figure*}

Figures \ref{GCN 45min} and \ref{GAT 45min} show the evaluation of the accuracy performance of the sparse training for different model sparsity. We observe that the error is generally stabilized until the sparsity reaches 85\% or even 90\%.
The turning point (marked as red circles) appears When the sparsity reaches about 85 percent. Therefore, as long as we keep the sparsity below 85\%, there is negligible significant impact on the accuracy of the training model, while we could introduce significant computational reduction and memory footprint reduction in computer system.

\subsubsection{Training FLOPs Evaluation}

We use floating point operations per second (FLOPs) to compare the number of operations between the sparse training version and dense training version (i.e., 0\% sparsity) for both GCN-STGT and GAT-STGT.
The ratio of FLOPs of the sparse model to the dense model has the following relationship:
\begin{align}
    \frac{FLOPs_{sparse}}{FLOPs_{dense}} = \frac{3\Delta T - 3d\Delta T - 2d + 3}{3(\Delta T + 1)}
\end{align}
where $\Delta T$ is the drop-and-grow frequency, and $d$ is the sparsity rate.

\begin{figure}
\centering
\includegraphics[scale=0.5]{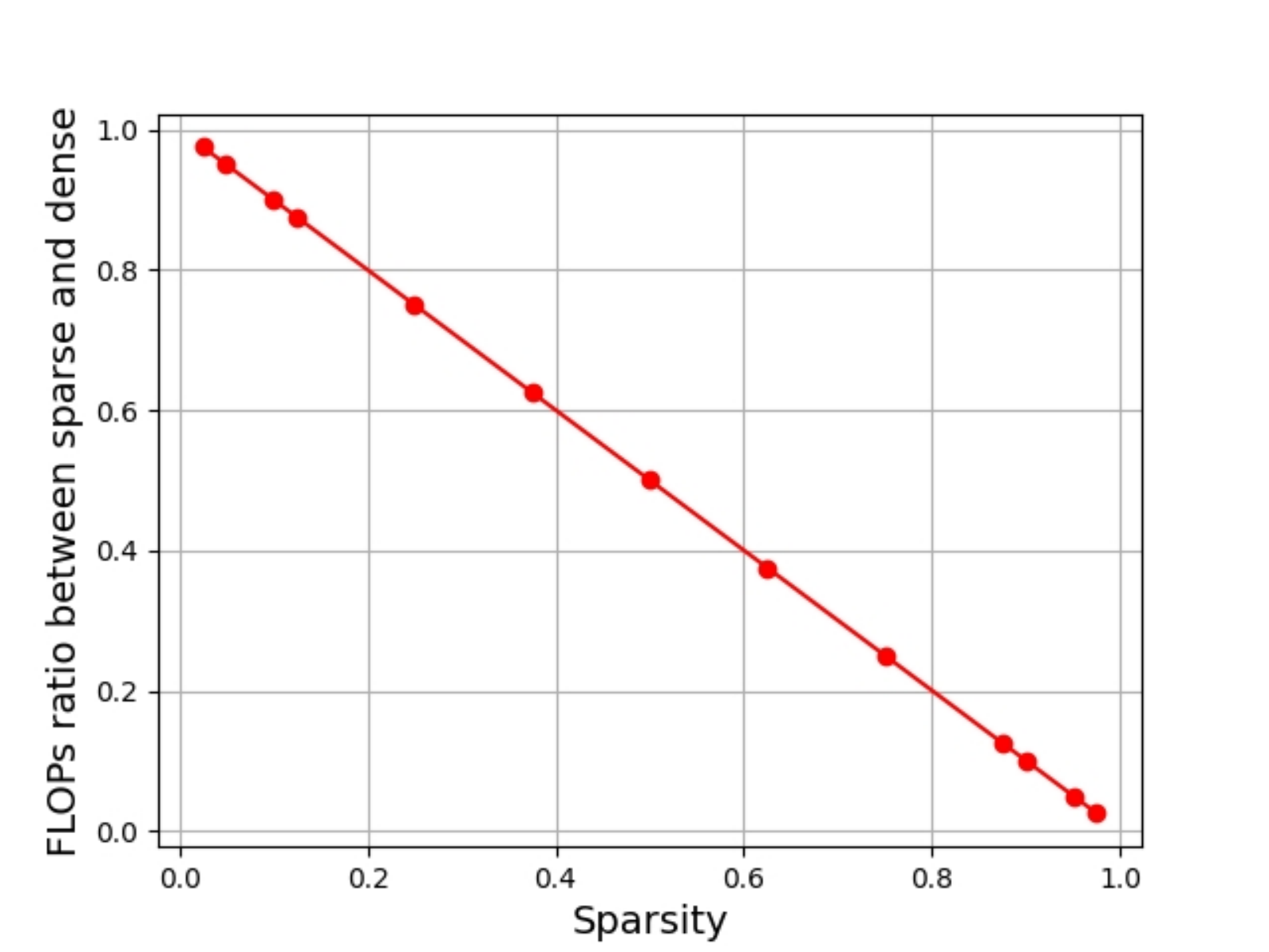}
\caption{The relationship between sparsity and  FLOPs ratio for sparse training.} \label{fig.3} 
\end{figure}

The calculation of FLOPS is shown below:
\begin{gather} 
\intertext{Let:}\Theta^{flop}_{M_{1 \times j} , M_{j \times 1}} = \xi
\intertext{Where $\Theta^{flop}_{M_{1 \times j}, M_{j \times 1}}$ means calculating the asymptotic bounds for the FLOPs of multiplication between matrix $M_{1 \times j}$ and matrix $M_{j \times 1}$}
\intertext{So:}\Theta^{flop}_{M_{i\times j}, M_{j \times k}} = ik\xi
\intertext{For simple fully connected layers:}\Theta^{flop}_{\textit{fully connected}} = IO \xi
\intertext{where $I$ is the input dimensionality and $O$ is the output dimensionality.}
\intertext{For convolutional kernels:}\Theta^{flop}_{\textit{convolutional}} = L_hL_wC_{in}C_{out}K^{2}\xi
\intertext{where $L_h$, $L_w$ and $C_{in}$ are height, width and number of channels of the input feature map, K is the kernel width (assumed to be symmetric), and $C_{out}$ is the number of output channels}
\intertext{Training a neural network mainly consist of three steps: calculate loss, calculate gradient, calculate weights. In the estimation process, it is assumed that all three require the same amount of FLOPs. Thus, for sparse training: the total amount is 3$f_s$, where the $f_s$ is the FLOPs for one step in sparse training. For dense training : the total amount is 3$f_d$, where the $f_d$ is the FLOPs for one step in dense training. In our model: the total FLOPs is:}
FLOPs = \frac{3f_s\Delta T + 2f_s + f_d}{\Delta T + 1}
\intertext{We need to calculate the dense gradients for updating connections every $\Delta T$ iteration.}
\intertext{When there are more fully connected layer nodes and a larger number of layers:}
\Theta^{flop}_{\textit{model}} =\sum_{i=1}^{L} I_{i}O_{i} \xi
\intertext{where $i$ means the $i$-th layer}
\intertext{For the sparse model:}
\Theta^{flop}_{\textit{model}} =\sum_{i=1}^{L} I_{i}O_{i} \xi (1-d)
\intertext{where $d$ denoted the sparsity rate}
\intertext{Thus the connection between $f_d$ and $f_s$ approximately equal to}
f_s = f_d \times (1-d)
\intertext{Hence:}
FLOPs =f_d \frac{3\Delta T - 3d\Delta T - 2d + 3}{\Delta T + 1}
\end{gather}
The ratio of FLOPs of the sparse models with respect to different sparsity is shown in Fig. \ref{fig.3}. Under a certain update frequency (1,000 iterations/time in our model), the sparsity is linearly related to FLOPS. According to Fig.~\ref{GCN 45min} and Fig.~\ref{GAT 45min}, our sparse training method could stabilize the prediction accuracy when sparsity is up to 90\%. Therefore, we could bring 10$\times$ FLOPs reduction throughout the training process, when the training epochs are the same. This can increase the speed of operations by nearly ten times.

\section{Conclusion}

In this work, to mitigate the long-term traffic prediction challenge, we integrate the spatio-temporal dependencies in the transportation network from network modeling, together with the GCN and GAT. To further tackle the dramatic computation and memory cost caused by the giant model size (i.e., number of weights) caused by multiple cascaded layers, we propose sparse training to mitigate the training cost, while preserving the prediction accuracy.  We test the proposed methods on a real large-scale transportation network data, Archived Data User Service (ADUS), from California Department of Transportation (Caltrans) Performance Measurement System (PeMS). Experimental results show that the proposed GCN-STGT and GAT-STGT methods achieve very low prediction error for short-, mid- and long-term prediction horizons. Using our sparse training,
we could train from scratch
with high sparsity (e.g., 90\%), equivalent to 10 times saving on computational cost using the same epochs as dense training,
and arrive at a model with very small accuracy loss compared with the original dense training.

\bibliographystyle{ieeetr}
\bibliography{ref}

\newpage
\section{Biography Section}

 

\vspace{-200pt}
\begin{IEEEbiography}[{\includegraphics[width=1in,height=1.25in,clip,keepaspectratio]{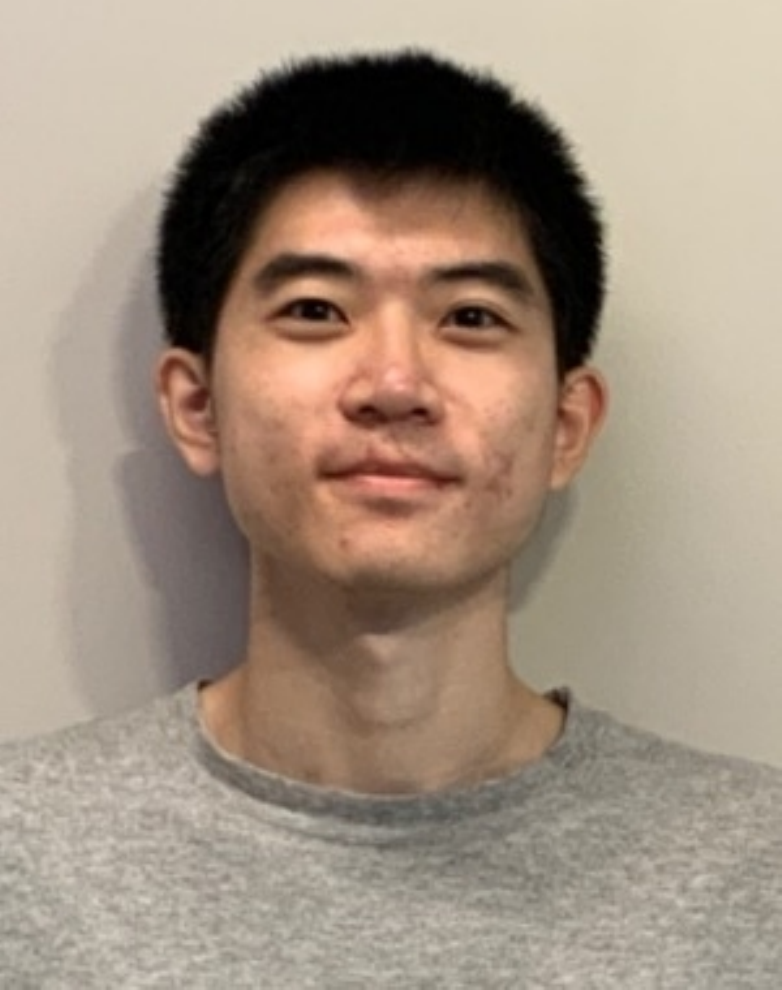}}]{Bin Lei}
is a Master student in the Department of Computer Science and Engineering at University of Connecticut, Storrs, Connecticut, USA.
His research interests include machine learning and algorithms.

\end{IEEEbiography}
\vspace{-200pt}
\begin{IEEEbiography}[{\includegraphics[width=1in,height=1.25in,clip,keepaspectratio]{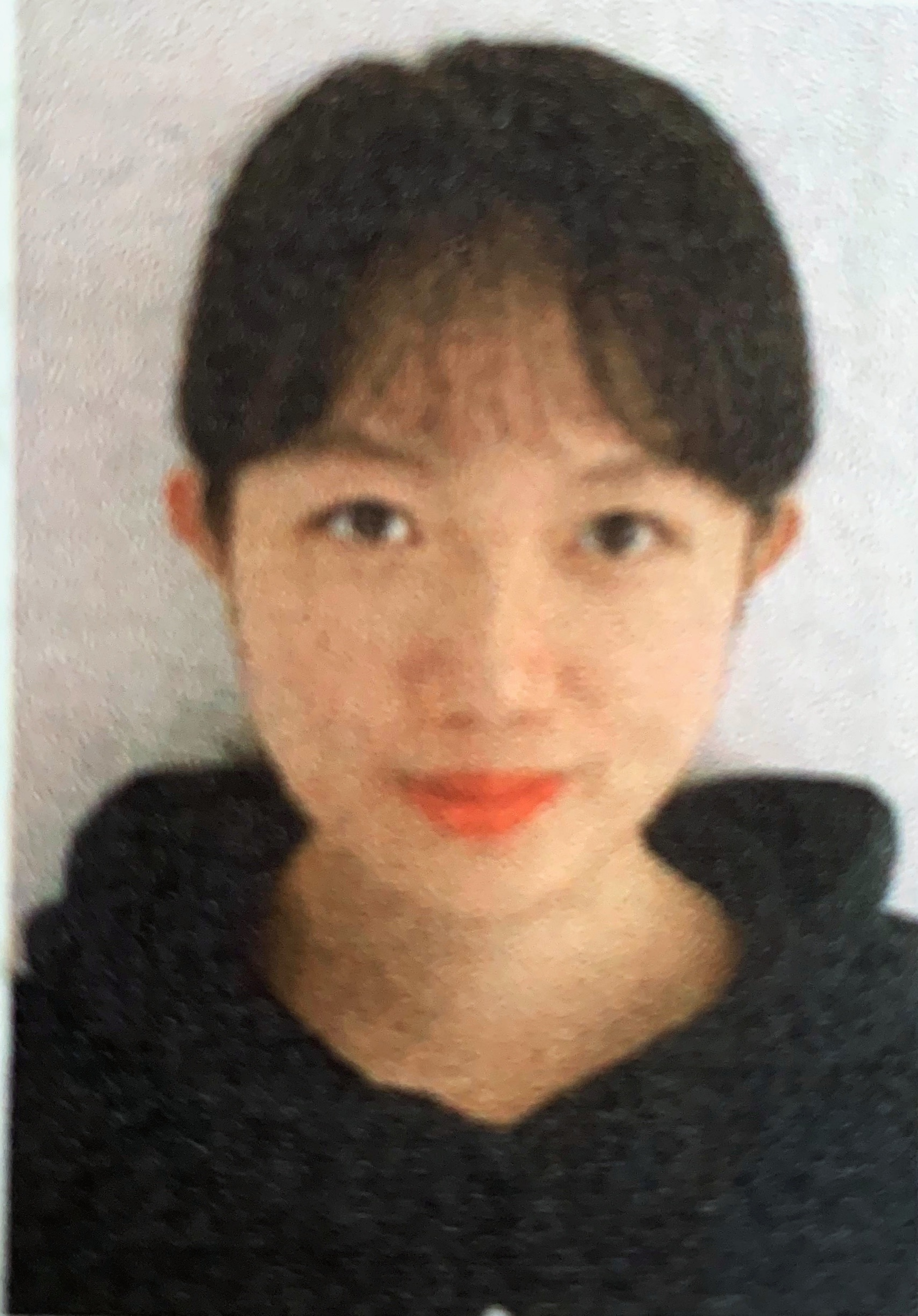}}]{Shaoyi Huang} (Student Member, IEEE)
is a third year Ph.D. student in the Department of Computer Science and Engineering at University of Connecticut, Storrs, Connecticut, USA.
Her research interests include deep learning, efficient AI, software / hardware co-design, natural language processing, computer vision.

\end{IEEEbiography}
\vspace{-200pt}

\begin{IEEEbiography}[{\includegraphics[width=1in,height=1.25in,clip,keepaspectratio]{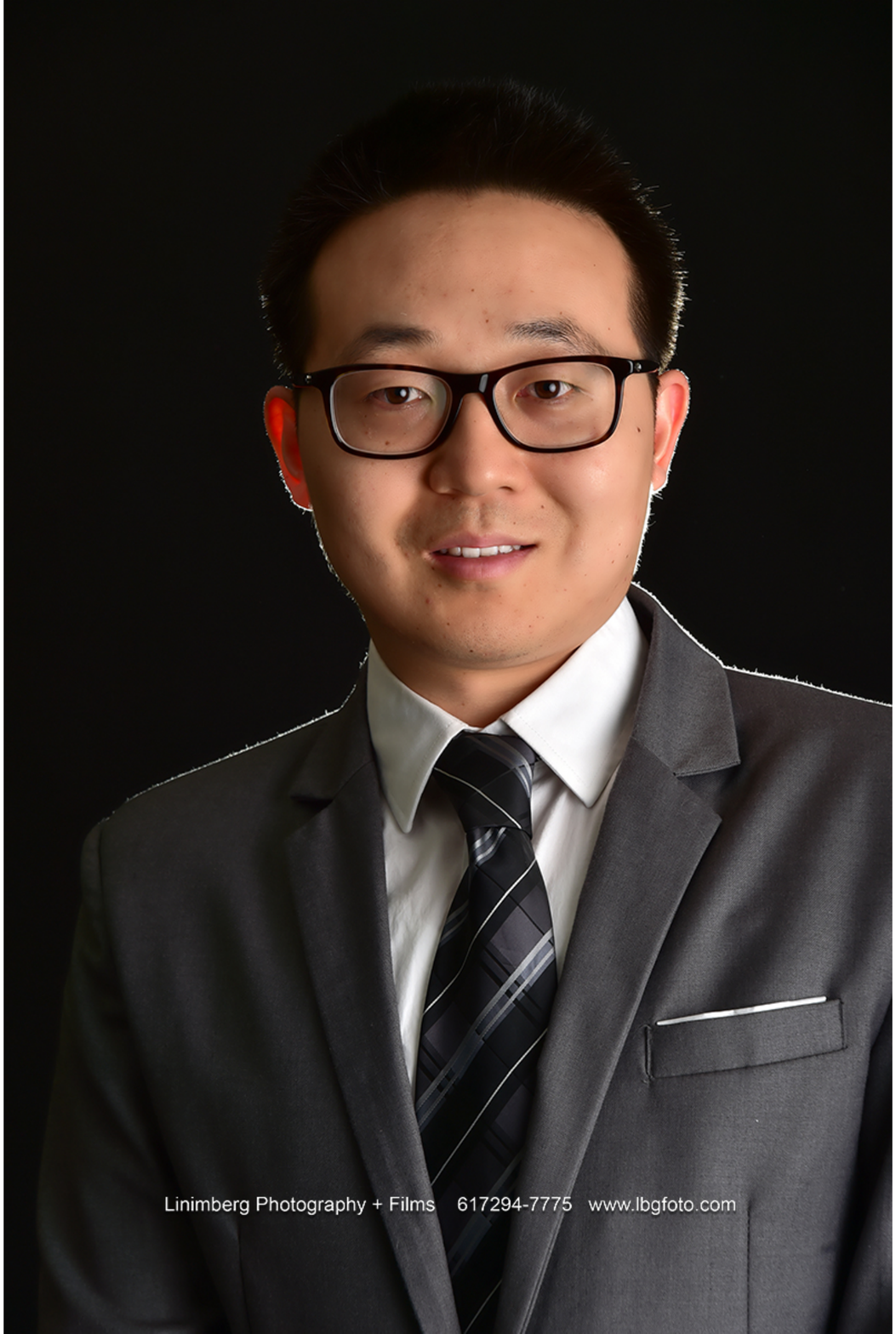}}]{Caiwen Ding}
is an assistant professor in the Department of Computer Science \& Engineering at the University of Connecticut. He received his Ph.D. degree from Northeastern University (NEU), Boston in 2019.
His interests include machine learning systems; computer architecture and heterogeneous computing (FPGAs/GPUs); computer vision, natural language processing; non-von Neumann computing and neuromorphic computing; privacy-preserving machine learning.


\end{IEEEbiography}
\vspace{-200pt}

\begin{IEEEbiography}[{\includegraphics[width=1in,height=1.25in,clip,keepaspectratio]{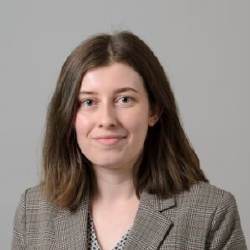}}]{Monika Filipovska}
is an assistant professor in the Department of Civil and Environmental Engineering at the University of Connecticut. She received her Ph.D. in transportation systems from Northwestern University. Her research focuses on predictive and prescriptive analytics for dynamic transportation networks and intelligent transportation systems, including applications of emerging vehicle and infrastructure technologies.


\end{IEEEbiography}




\end{document}